# Exploring Factors Affecting Pedestrian Crash Severity Using TabNet: A Deep Learning Approach

**Amir Rafe** (amir.rafe@usu.edu)
Graduate Research Assistant, Civil & Environmental Engineering, Utah State University

**Patrick A. Singleton** (patrick.singleton@usu.edu)
Associate Professor, Civil & Environmental Engineering, Utah State University

## Abstract

This study presents the first investigation of pedestrian crash severity using the TabNet model, a novel tabular deep learning method exceptionally suited for analyzing the tabular data inherent in transportation safety research. Through the application of TabNet to a comprehensive dataset from Utah covering the years 2010 to 2022, we uncover intricate factors contributing to pedestrian crash severity. The TabNet model, capitalizing on its compatibility with structured data, demonstrates remarkable predictive accuracy, eclipsing that of traditional models. It identifies critical variables—such as pedestrian age, involvement in left or right turns, lighting conditions, and alcohol consumption—which significantly influence crash outcomes. The utilization of SHapley Additive exPlanations (SHAP) enhances our ability to interpret the TabNet model's predictions, ensuring transparency and understandability in our deep learning approach. The insights derived from our analysis provide a valuable compass for transportation safety engineers and policymakers, enabling the identification of pivotal factors that affect pedestrian crash severity. Such knowledge is instrumental in formulating precise, data-driven interventions aimed at bolstering pedestrian safety across diverse urban and rural settings.

## Introduction

Pedestrian safety remains a critical challenge in traffic systems worldwide, with pedestrians often bearing the highest risk of traffic crashes. In 2021 alone, the National Highway Traffic Safety Administration (NHTSA) (*1*) reported 7,388 pedestrian fatalities in the United States, underscoring the need for improved safety measures. Various factors contribute to the severity of pedestrian crashes, with urban settings, intersections, and low-light conditions being predominant risk factors.

Data-driven analysis of crash reports is a key strategy for identifying factors that influence pedestrian crash severity. Recently, deep learning techniques have shown promise in this domain due to their ability to capture complex patterns from large volumes of data. This study harnesses the potential of TabNet, a state-of-the-art deep learning model designed for tabular data, which is prevalent in the field of transportation safety. TabNet's innovative architecture enables it to focus on the most relevant factors for crash severity prediction, thereby offering a powerful tool for traffic safety analysis.

Utilizing pedestrian crash data from Utah spanning 2010 to 2021, this study is the first to apply TabNet to pedestrian crash severity analysis. In conjunction with SHapley Additive exPlanations (SHAP), we interpret the model's predictions, providing insights into the significance of various contributing factors. This novel approach not only enhances model interpretability but also aids in developing targeted strategies to improve pedestrian safety. The ensuing sections will detail the methodology, present the findings, and discuss the implications of employing TabNet within this vital area of public safety.

## Literature Review

The severity of traffic incidents involving pedestrians is contingent upon a myriad of factors. A comprehensive review of relevant academic literature (*2*) reveals several key variables. These include the demographic characteristics of the pedestrian, with a particular emphasis on age and gender; the speed and type of the implicated vehicle; the details of the accident location and the timing of the incident; the presence of intoxicating substances in the pedestrian or driver; and the use of safety equipment such as helmets or high-visibility clothing. These elements collectively contribute to the understanding and assessment of pedestrian-related traffic incidents.

Numerous prediction models have been employed to investigate the impact of various factors on pedestrian crash severity. These models encompass statistical techniques, such as negative binomial models (*3*), logistic regression models (*4*), ordered probit models (*5, 6*), and structural equation modeling (*7*). Machine learning (ML) models, including random forest, AdaBoost (*8*), XGBoost (*9*), decision trees, k-nearest neighbor, and ensemble models (*10*) have also been utilized. Additionally, deep learning (DL) models, like deep neural networks (DNN) (*11*), have been explored for pedestrian crash severity analysis. To gain a better understanding of the application of these techniques, Table 1 presents an overview of the advantages and limitations of these methods used in pedestrian crash severity analysis.

**TABLE 1 Summary of benefits and limitations of various techniques for pedestrian crash severity analysis**

| Techniques | Benefits | Limitations |
|---|---|---|
| Statistical methods | - *Interpretability:* Statistical models, such as logistic regression and negative binomial models, offer greater interpretability and understanding compared to ML and DL models (*7, 12*).<br>- *Simplicity:* Statistical models are generally simpler and require fewer computational resources than ML and DL models (*12*).<br>- *Well-established techniques:* Statistical methods have a long history of use and research, making them reliable and well-established for analyzing crash severity (*7*). | - *Linearity assumptions:* Some statistical models, like logistic regression, may assume a linear relationship between predictors and the outcome, which could be limited in capturing more complex real-world scenarios (*12*).<br>- *Limited predictive power:* Statistical models might have lower predictive accuracy compared to ML and DL models, especially when handling intricate and non-linear relationships between variables (*12*). |
| ML and DL methods | - *Higher predictive accuracy:* ML and DL methods can achieve superior predictive accuracy compared to statistical models, particularly when handling complex and non-linear relationships between variables or when dealing with large and complex datasets (*11, 13*).<br>- *Feature importance:* ML models can effectively identify significant features (explanatory variables) and their relationships with crash severity, providing valuable insights that might be more challenging to extract from statistical models (*13*). | - *Interpretability:* ML and DL models can be more challenging to interpret and comprehend than statistical models, which may hinder the ability to explain the relationships between variables and crash severity (*12*).<br>- *Overfitting:* ML and DL models may be susceptible to overfitting, particularly when dealing with many features or a small dataset. This can lead to reduced generalizability and accuracy on unseen data (*11, 13*).<br>- *Computational resources:* DL models typically demand more computational resources and longer training times in comparison to statistical and ML models (*11*). |

While TabNet (*14*) (a DL technique designed for tabular data analysis, capable of handling both numerical and categorical variables) has been used in crash severity analysis before, our study is novel in terms of applying TabNet specifically to pedestrian crash severity analysis. Prior work by Sattar et al. (*15*) utilized TabNet for modeling injury severity in motor vehicle crashes using different ML approaches. However, their study did not focus on pedestrian-related crashes, and they did not propose the TabNet interpretation results. Therefore, our study also contributes by introducing the interpretation of TabNet results using SHAP, a framework previously employed for interpreting DNN models in crash injury severity analysis by Kang et al. (*11*), and for XGBoost models in similar studies by Chang et al. (*16*) and Li (*17*). By incorporating SHAP, we aim to provide deeper insights into the factors influencing pedestrian crash severity predictions using the TabNet model.

## Data and Method

### Data and Variables

In our research, we leveraged crash data (*18*) to explore the determinants of pedestrian crash severities in Utah over the period from 2010 to 2021. The severity of pedestrian crashes in our study was gauged using the KABCO scale. This scale classifies crashes into several categories: fatal, suspected serious injury, suspected minor injury, possible injury, and no injury or property-damage-only (PDO). To visually represent this data, Figure 1 showcases the spatial distribution of these crashes. Additionally, it includes a heatmap that accentuates the locations of fatal crashes within the dataset.

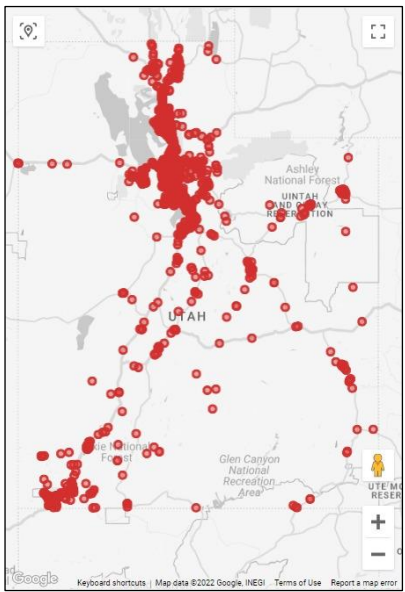 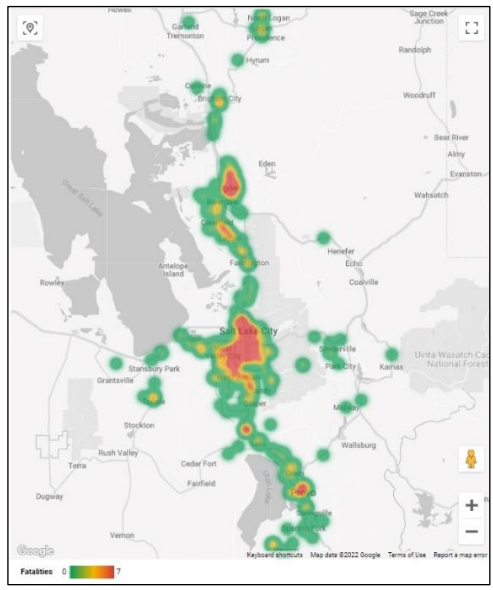

(a) Pedestrian crashes dispersion            (b) Heatmap of pedestrian fatalities

**Figure 1 The spatial configuration of pedestrian crashes**

In this study, we examined 8,812 pedestrian crash incidents, analyzing the impact of 29 different variables, as detailed in Table 2. The breakdown of crash severities was as follows: fatal crashes comprised 5%, serious injuries 15%, minor injuries 44%, possible injuries 30%, and no injuries or property-damage-only (PDO) accounted for 6%. Notable insights from the data include a higher incidence of injury among male pedestrians and an increased rate of fatalities in the 30 to 59 age group. Factors like DUI (Driving Under the Influence) and crashes involving older drivers contributed to 13% and 11% of pedestrian fatalities, respectively. A significant majority of these crashes occurred on arterial roads (52%) and predominantly in urban areas (97%). Intersections emerged as common sites for pedestrian crashes, accounting for 61% of the total, with nearly 3% of these being fatal. The study also found that left-turn and right-turn accidents occurred at similar rates. Regarding lighting conditions, 60% of crashes happened in daylight, while dark conditions without lighting were present in 37% of fatal crashes.

**TABLE 2 Descriptive statistics of the variables**

| Characteristics | Class | Total | Fatal | Serious injury | Minor injury | Possible injury | No injury / PDO |
|---|---|---|---|---|---|---|---|
| Pedestrian crashes | | 8812 (0%) | 476 (5%) | 1363 (15%) | 3856 (44%) | 2624 (30%) | 493 (6%) |
| Sex | Male | 5282 (60%) | 309 (65%) | 849 (62%) | 2261 (59%) | 1514 (58%) | 349 (71%) |
| | Female | 3530 (40%) | 167 (35%) | 514 (38%) | 1595 (41%) | 1110 (42%) | 144 (29%) |
| Age group | 0 to 9 | 826 (9%) | 33 (7%) | 118 (9%) | 391 (10%) | 240 (9%) | 44 (9%) |
| | 10 to 29 | 4039 (46%) | 119 (25%) | 556 (41%) | 1840 (48%) | 1280 (49%) | 244 (49%) |
| | 30 to 59 | 3009 (34%) | 202 (42%) | 509 (37%) | 1265 (33%) | 866 (33%) | 167 (34%) |
| | > 59 | 938 (11%) | 122 (26%) | 180 (13%) | 360 (9%) | 238 (9%) | 38 (8%) |
| Aggressive driving | No | 8703 (99%) | 472 (99%) | 1332 (98%) | 3812 (99%) | 2601 (99%) | 486 (99%) |
| | Yes | 109 (1%) | 4 (1%) | 31 (2%) | 44 (1%) | 23 (1%) | 7 (1%) |
| Alcohol-drug test result | Both-Positive | 11 (0%) | 11 (2%) | 0 (0%) | 0 (0%) | 0 (0%) | 0 (0%) |
| | Drug-Positive | 34 (0%) | 34 (7%) | 0 (0%) | 0 (0%) | 0 (0%) | 0 (0%) |
| | Alcohol-Positive | 15 (0%) | 13 (3%) | 2 (0%) | 0 (0%) | 0 (0%) | 0 (0%) |
| | Negative | 9 (0%) | 9 (2%) | 0 (0%) | 0 (0%) | 0 (0%) | 0 (0%) |
| | Not related | 8743 (99%) | 409 (86%) | 1361 (100%) | 3856 (100%) | 2624 (100%) | 493 (100%) |
| DUI | No | 8586 (97%) | 413 (87%) | 1305 (96%) | 3786 (98%) | 2598 (99%) | 484 (98%) |
| | Yes | 226 (3%) | 63 (13%) | 58 (4%) | 70 (2%) | 26 (1%) | 9 (2%) |
| Distracted driving | No | 8105 (92%) | 430 (90%) | 1218 (89%) | 3554 (92%) | 2449 (93%) | 454 (92%) |
| | Yes | 707 (8%) | 46 (10%) | 145 (11%) | 302 (8%) | 175 (7%) | 39 (8%) |
| Drowsy driving | No | 8773 (100%) | 463 (97%) | 1354 (99%) | 3846 (100%) | 2620 (100%) | 490 (99%) |
| | Yes | 39 (0%) | 13 (3%) | 9 (1%) | 10 (0%) | 4 (0%) | 3 (1%) |
| Older driver involved | No | 7896 (90%) | 423 (89%) | 1223 (90%) | 3446 (89%) | 2366 (90%) | 438 (89%) |
| | Yes | 916 (10%) | 53 (11%) | 140 (10%) | 410 (11%) | 258 (10%) | 55 (11%) |
| Teenage driver involved | No | 7966 (90%) | 432 (91%) | 1205 (88%) | 3496 (91%) | 2390 (91%) | 443 (90%) |
| | Yes | 846 (10%) | 44 (9%) | 158 (12%) | 360 (9%) | 234 (9%) | 50 (10%) |
| Holiday | No | 7745 (88%) | 397 (83%) | 1185 (87%) | 3402 (88%) | 2324 (89%) | 437 (89%) |
| | Yes | 1067 (12%) | 79 (17%) | 178 (13%) | 454 (12%) | 300 (11%) | 56 (11%) |
| Right-turn involved | No | 7118 (81%) | 464 (97%) | 1254 (92%) | 3113 (81%) | 1912 (73%) | 375 (76%) |
| | Yes | 1694 (19%) | 12 (3%) | 109 (8%) | 743 (19%) | 712 (27%) | 118 (24%) |
| Intersection involved | Yes | 5361 (61%) | 136 (29%) | 718 (53%) | 2471 (64%) | 1766 (67%) | 270 (55%) |
| | No | 3451 (39%) | 340 (71%) | 645 (47%) | 1385 (36%) | 858 (33%) | 223 (45%) |
| Left-turn involved | No | 7079 (80%) | 441 (93%) | 1144 (84%) | 3016 (78%) | 2051 (78%) | 427 (87%) |
| | Yes | 1733 (20%) | 35 (7%) | 219 (16%) | 840 (22%) | 573 (22%) | 66 (13%) |
| Overturn rollover | No | 8785 (100%) | 474 (100%) | 1358 (100%) | 3842 (100%) | 2620 (100%) | 491 (100%) |
| | Yes | 27 (0%) | 2 (0%) | 5 (0%) | 14 (0%) | 4 (0%) | 2 (0%) |
| Domestic animal involved | No | 8793 (100%) | 469 (99%) | 1363 (100%) | 3847 (100%) | 2622 (100%) | 492 (100%) |
| | Yes | 19 (0%) | 7 (1%) | 0 (0%) | 9 (0%) | 2 (0%) | 1 (0%) |
| Commercial vehicle involved | No | 8559 (97%) | 440 (92%) | 1302 (96%) | 3772 (98%) | 2581 (98%) | 464 (94%) |
| | Yes | 253 (3%) | 36 (8%) | 61 (4%) | 84 (2%) | 43 (2%) | 29 (6%) |
| Heavy truck involved | No | 8534 (97%) | 440 (92%) | 1297 (95%) | 3760 (98%) | 2577 (98%) | 460 (93%) |
| | Yes | 278 (3%) | 36 (8%) | 66 (5%) | 96 (2%) | 47 (2%) | 33 (7%) |
| Transit vehicle involved | No | 8732 (99%) | 470 (99%) | 1349 (99%) | 3823 (99%) | 2606 (99%) | 484 (98%) |
| | Yes | 80 (1%) | 6 (1%) | 14 (1%) | 33 (1%) | 18 (1%) | 9 (2%) |
| Work zone involved | No | 8425 (96%) | 447 (94%) | 1298 (95%) | 3705 (96%) | 2503 (95%) | 472 (96%) |
| | Yes | 387 (4%) | 29 (6%) | 65 (5%) | 151 (4%) | 121 (5%) | 21 (4%) |
| Wrong way driving | No | 8784 (100%) | 473 (99%) | 1359 (100%) | 3841 (100%) | 2620 (100%) | 491 (100%) |
| | Yes | 28 (0%) | 3 (1%) | 4 (0%) | 15 (0%) | 4 (0%) | 2 (0%) |
| Road type | Urban | 8548 (97%) | 419 (88%) | 1296 (95%) | 3784 (98%) | 2583 (98%) | 466 (95%) |
| | Rural | 264 (3%) | 57 (12%) | 67 (5%) | 72 (2%) | 41 (2%) | 27 (5%) |
| Functional class | Local | 2651 (30%) | 71 (15%) | 352 (26%) | 1256 (33%) | 847 (32%) | 125 (25%) |
| | Collector | 1578 (18%) | 71 (15%) | 232 (17%) | 703 (18%) | 487 (19%) | 85 (17%) |
| | Arterial | 4583 (52%) | 334 (70%) | 779 (57%) | 1897 (49%) | 1290 (49%) | 283 (57%) |
| Roadway surface is dry | Yes | 7607 (86%) | 409 (86%) | 1181 (87%) | 3312 (86%) | 2273 (87%) | 432 (88%) |
| | No | 1205 (14%) | 67 (14%) | 182 (13%) | 544 (14%) | 351 (13%) | 61 (12%) |
| Lighting condition | Dark-Not lighted | 1167 (13%) | 176 (37%) | 285 (21%) | 401 (10%) | 255 (10%) | 50 (10%) |
| | Dark-Lighted | 1912 (22%) | 138 (29%) | 332 (24%) | 824 (21%) | 542 (21%) | 76 (15%) |
| | Daylight | 5292 (60%) | 141 (30%) | 678 (50%) | 2409 (62%) | 1725 (66%) | 339 (69%) |
| | Dusk | 244 (3%) | 10 (2%) | 40 (3%) | 115 (3%) | 59 (2%) | 20 (4%) |
| | Dawn | 197 (2%) | 11 (2%) | 28 (2%) | 107 (3%) | 43 (2%) | 8 (2%) |

**TABLE 2 (Continued)**

| Characteristics | Class | Total | Fatal | Serious injury | Minor injury | Possible injury | No injury / PDO |
|---|---|---|---|---|---|---|---|
| Weather condition | Clear | 6758 (77%) | 355 (75%) | 1068 (78%) | 2925 (76%) | 2030 (77%) | 380 (77%) |
| | Cloudy | 1214 (14%) | 69 (14%) | 176 (13%) | 543 (14%) | 353 (13%) | 73 (15%) |
| | Rain | 509 (6%) | 31 (7%) | 80 (6%) | 220 (6%) | 153 (6%) | 25 (5%) |
| | Fog, Smog | 25 (0%) | 3 (1%) | 4 (0%) | 9 (0%) | 8 (0%) | 1 (0%) |
| | Snowing | 213 (2%) | 10 (2%) | 26 (2%) | 112 (3%) | 53 (2%) | 12 (2%) |
| | Others | 93 (1%) | 8 (2%) | 9 (1%) | 47 (1%) | 27 (1%) | 2 (0%) |
| Vertical alignment | Level | 6891 (78%) | 360 (76%) | 1108 (81%) | 3005 (78%) | 2042 (78%) | 376 (76%) |
| | Uphill | 61 (1%) | 3 (1%) | 7 (1%) | 33 (1%) | 14 (1%) | 4 (1%) |
| | Downhill | 50 (1%) | 2 (0%) | 12 (1%) | 28 (1%) | 8 (0%) | 0 (0%) |
| | Others | 1810 (21%) | 111 (23%) | 236 (17%) | 790 (20%) | 560 (21%) | 113 (23%) |

In the development of our TabNet models, we adhered to the categorization outlined in Table 2. To ensure a uniform encoding of the dataset, we assigned numerical values to categorical data. For instance, we designated "Yes" as 1 and "No" as 0; "Male" received a value of 1, while "Female" was assigned 0; similarly, "Rural" was encoded as 1 and "Urban" as 0. Other categories were numerically encoded following their sequential arrangement in Table 2, starting from 1 and increasing. Furthermore, we treated age as a continuous variable, rather than categorizing it into different age groups.

**Method**

In this study, we utilized the TabNet methodology to delve into the effects of various explanatory variables on pedestrian crash injury outcomes. TabNet, a model tailored for tabular data, is celebrated for its robust performance and interpretability, initially developed by the team at Google Cloud AI (*14*). It ingeniously merges the capabilities of deep learning models with feature selection techniques, adept at processing both numerical and categorical data. TabNet's core functionality lies in its use of sequential attention. This feature allows the model to selectively focus on different explanatory variables (EVs) at each decision-making step, thereby enhancing its interpretability. Figure 2 in our study depicts the specific structure of the TabNet model as applied here, highlighting its architecture over two steps. Within this framework, the EV transformer plays a crucial role in refining input data, which helps in better understanding the interplay between EVs and crash severity levels. Concurrently, the attentive transformer assesses the significance of each EV during each decision step. It creates a mask to emphasize the most influential predictors, enabling the model to dynamically concentrate on pertinent factors such as weather conditions, alcohol involvement, among other EVs. This approach not only bolsters the model's focus but also significantly augments the accuracy of its predictions.

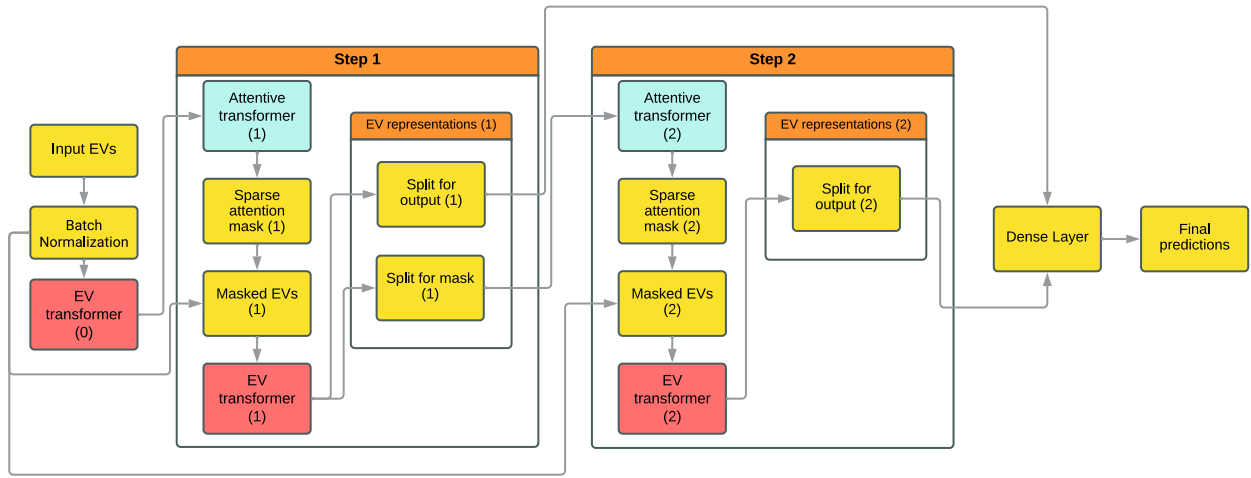

**Figure 2 The structure of the TabNet model for predicting crash severity levels using various EVs.**

When applying the TabNet model to predict pedestrian crash severity, we implemented several steps to optimize its performance and accuracy. To counter the class imbalance in our dataset, we used the Synthetic Minority Over-sampling Technique (SMOTE) (*19*), enhancing the model's proficiency in predicting less-represented classes. The model's hyperparameters were fine-tuned with the help of Optuna (*20*), a framework specialized in hyperparameter optimization, to achieve the best possible configuration tailored to our specific dataset. Additionally, to prevent overfitting and improve the model's ability to generalize, we conducted multiple training iterations on varied subsets of data through bootstrapping. We evaluated the model's effectiveness using a range of metrics, including accuracy, precision, recall, and the F1-score, to ensure a comprehensive assessment of its performance. The calculations for these metrics are represented by Eq. 1 to 4:

$$accuracy = \frac{True\ Positive + True\ Negative}{True\ Positive + True\ Negative + False\ Positive + False\ Negative} \tag{1}$$

$$precision = \frac{True\ Positive}{True\ Positive + False\ Positive} \tag{2}$$

$$recall = \frac{True\ Positive}{True\ Positive + False\ Negative} \tag{3}$$

$$F1\ score = \frac{2 \times (precision \times recall)}{precision + recall} \tag{4}$$

For interpreting the results of our TabNet model, we employed SHapley Additive exPlanations (SHAP) (*21, 22*). SHAP assigns an importance value to each explanatory variable (EV) for a given prediction, making the model's output more understandable in terms of the input EVs. Drawing from cooperative game theory, SHAP values distribute the prediction output (crash severity) among the EVs based on their contribution. If we denote f(x) as the prediction for a specific instance x and $E[f(X)]$ as the expected prediction for the model, which is calculated as the average prediction over the training dataset, the additive EV attribution can be calculated as follows:

$$f(x) - E[f(X)] = \sum_{i=1}^{N} \varphi_i \tag{5}$$

Furthermore, the importance value assigned to each EV or the Shapley value for the i-th EV $\varphi_i$ can be calculated as follows:

$$\varphi_i = \sum_{S \subseteq N\{i\}} \left[ \frac{|S|!(|N| - |S| - 1)!}{|N|!} \right] (f_i(S \cup \{i\}) - f_i(s)) \tag{6}$$

where $N$ is the set of all EVs, $S$ is a subset of $N$ that includes the i-th EV, |S| is the size of S, and $f_i$ is a version of where only the EVs in S and i (if it's included) are used. For this analysis, the SHAP python package (*23*) was utilized to determine the importance of EVs in the TabNet model.

## Model Results

For evaluating the various models, we partitioned our data, dedicating 80% for training purposes and reserving the final 20% for testing and evaluation. In the case of the TabNet model, we employed the SHAP method to discern the significance of each explanatory variable (EV) across different crash severity classes. This approach and its findings are illustrated in Figure 3, providing a clear visual representation of how each EV influences the model's predictions for each severity level.

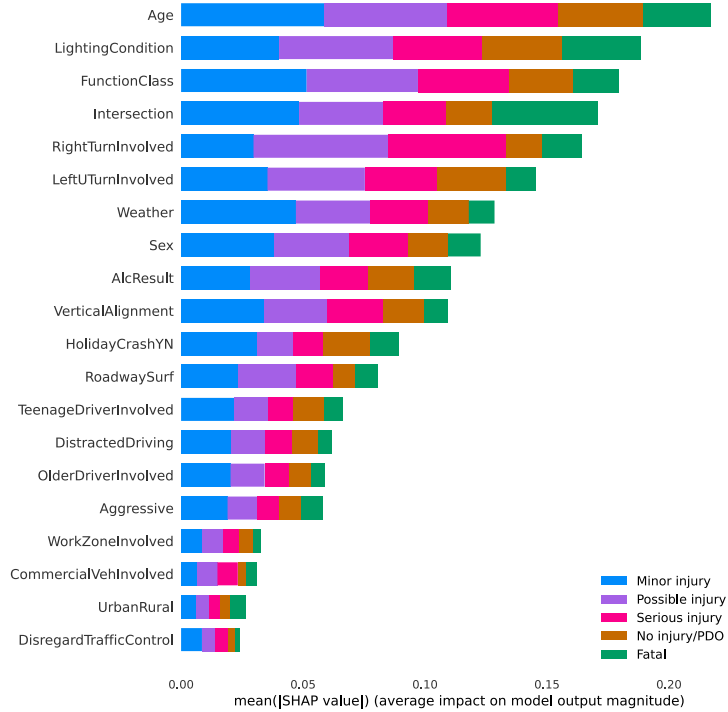

**Figure 3 The importance of each EV for each crash severity class in TabNet model**

To enhance the precision of the TabNet model, we meticulously adjusted its hyperparameters through a combination of GridSearchCV (*24*) and the Optuna optimization technique. The specific values and methods utilized for this fine-tuning process are comprehensively listed in Table 3.

**TABLE 3 Optimum hyperparameters of the TabNet models in this study**

| Model | Hyperparameter | Value/Method |
|---|---|---|
| TabNet | - Dimension of the prediction layer | 53 |
| | - Dimension of the attention layer | 58 |
| | - Number of decision steps | 1 |
| | - Sparsity regularization | 0.023989318 |
| | - Entmax* temperature** (gamma) | 1.952667709 |
| | - Number of independent GLU*** layers | 8 |
| | - Number of shared GLU layers across decision steps | 6 |
| | - Momentum in batch normalization | 0.3 |
| | - Gradient clipping for each parameter | 2 |
| | - Optimizer Function | Adam |
| | - Learning Rate (lr) | 0.007566832 |
| | - Mask Type | Entmax |

* It is a combination of "Maximum" and "Entropy," which signifies the objective of maximizing entropy while adhering to specific constraints. ** It is a hyperparameter that controls the sharpness of the probability distribution. *** Gated linear unit

For assessing the TabNet model's efficacy, we employed a suite of evaluation metrics including precision, recall, F-1 score, and overall accuracy. The outcomes derived from these metrics, offering insights into the model's performance, are detailed in Table 4.

**TABLE 4 Performance evaluation metrics**

| Crash severity class | Evaluation metrics | | |
|---|---|---|---|
| | **Precision** | **Recall** | **F1 score** |
| Fatal | 0.910 | 0.950 | 0.930 |
| Serious injury | 0.860 | 0.860 | 0.860 |
| Minor injury | 0.927 | 0.980 | 0.950 |
| Possible injury | 0.960 | 0.970 | 0.959 |
| No injury/PDO | 0.948 | 0.916 | 0.927 |
| **Accuracy** | 0.959 | | |

## Model Interpretation and Discussion

When we examined the performance metrics, as detailed in Table 4, the TabNet model distinguished itself with its precise predictions of pedestrian crash severity. It demonstrated particular strength in predicting minor and possible injury outcomes, as evidenced by its F1-score in these categories. To maintain the integrity of the TabNet model and to address the risk of overfitting due to its notable accuracy, we employed a range of methods. These included cross-validation, regularization parameters, an early stopping mechanism, the use of SMOTE, and training with diverse bootstrap samples. These strategies collectively improved both the performance and dependability of the model in our analysis.

The TabNet model's findings highlight pedestrian age, lighting conditions, and road functional class as key explanatory variables (EVs) in predicting crash severity. Figure 3 elucidates these influential EVs. To further understand the model, Figure 4 offers a SHAP summary plot that correlates EV features with crash severity classes. In this plot, each row represents an EV, with the color of the dots indicating the EV's value (red for high, blue for low), and their horizontal position indicating how the EV influences the probability of a higher severity outcome. The clustering of dots indicates a strong correlation between the feature and the prediction, with the spread showing the EV's impact and dot dispersion highlighting variation due to interactions with other EVs.

From the dot plot, certain features stand out for increasing the likelihood of fatal outcomes. Figure 4(a) shows that alcohol or drug consumption by pedestrians, and crashes in urban areas, are linked with higher chances of fatal severity. In serious injury cases (Figure 4(b)), the involvement of commercial vehicles and heavy trucks is a critical factor, though other EVs show variable impacts, suggesting complex interplays within the model. In minor injury cases, factors like disregard for traffic control, distracted driving, and holiday-period crashes are predictors, while right- and left-turns, work zones, and dry road conditions are more associated with possible injuries. Contrarily, adverse weather and commercial vehicle involvement reduce the odds of possible injuries. For the no injury/PDO category, holidays, left-turns, and the involvement of older or teenage drivers are inversely related to severity. These insights, provided by the SHAP analysis in Figure 4, highlight the nuanced interplay of various factors in pedestrian crash severity outcomes, as captured by the TabNet model.

To navigate the complexity of the dot summary plot, especially for intricate categories like serious injury, and to delve deeper into how each explanatory variable (EV) contributes to the final prediction, we employed SHAP force plots, exemplified in Figure 5 using observation #631 from our dataset. These force plots visually depict the influence of each EV on the model's prediction, starting from the base value (the average prediction) and culminating in the specific outcome for an observation. Here, the impact of each EV is shown as a horizontal force, indicating its effect in either increasing or decreasing the prediction likelihood.

In Figure 5(a), focusing on observation #631, the TabNet model shows a tendency to classify this case as fatal (f(x)=1). Factors like the alcohol result, presence at an intersection, age, involvement in a left-turn, and roadway surface type all point towards a fatal outcome. Conversely, the lighting condition applies a minor negative impact, but it's insufficient to outweigh the substantial positive influences from the other variables.

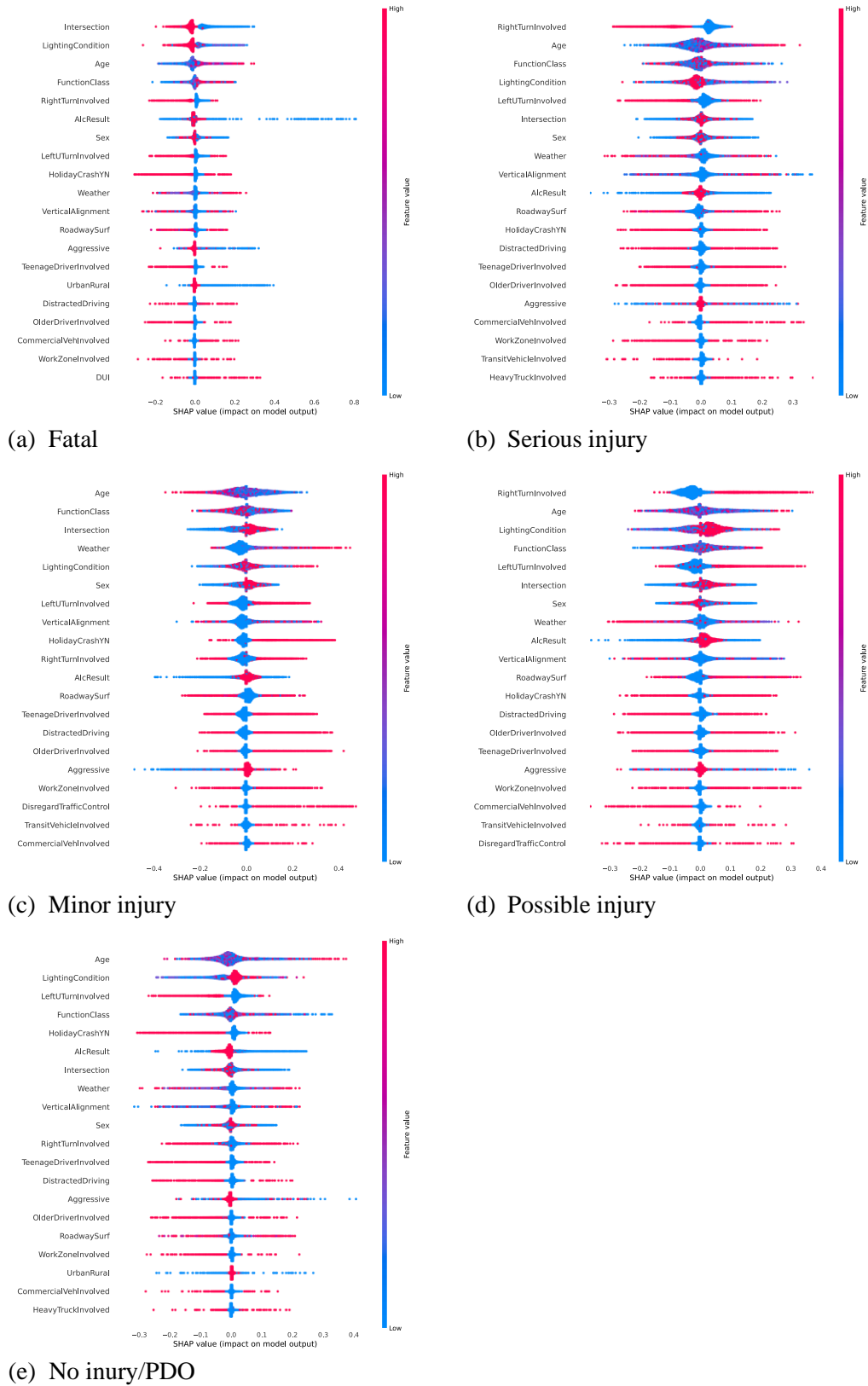

(a) Fatal

(b) Serious injury

(c) Minor injury

(d) Possible injury

(e) No inury/PDO

**Figure 4 The SHAP summary plot for each crash severity class in TabNet model**

Additionally, in Figure 5(b), the model predicts a non-serious injury outcome (f(x)=0). The base value, ranging between 0.15 and 0.20, acts as the starting point in the absence of specific information about this observation. A significant blue arrow indicates that the vertical alignment variable heavily influences the prediction towards f(x) = 0.00. Factors like age, left-turn involvement, alcohol result, and lighting condition also contribute negatively, albeit to a lesser extent. In contrast, the right-turn involvement (indicated by a pink arrow) partly mitigates but doesn't fully offset the negative influences. This pattern is representative across other categories as well. From this analysis, it is clear that the model accurately classified observation #631 as fatal. Among the influencing features, the most impactful was the alcohol result, underscoring its significance in determining crash severity in this instance.

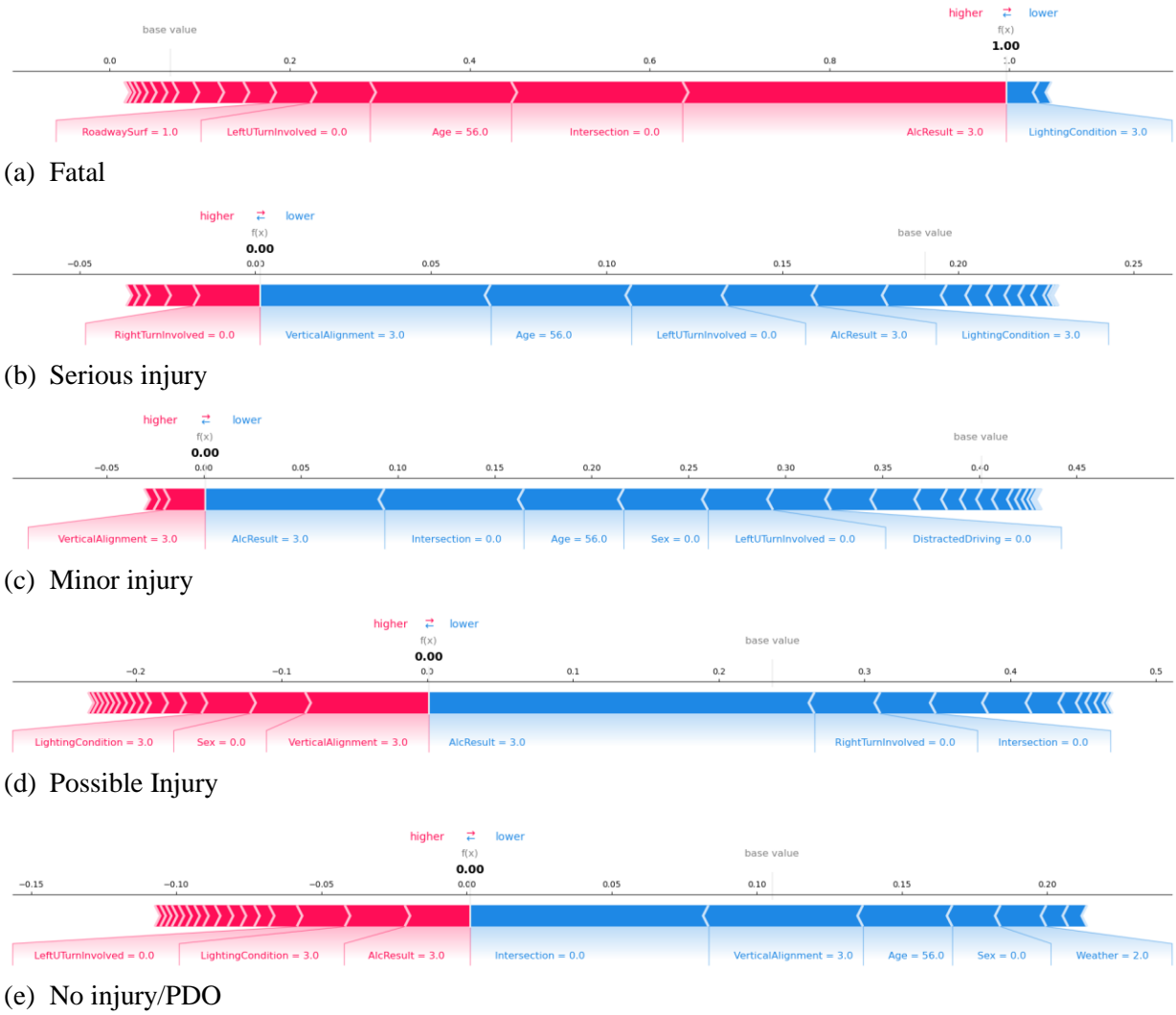

(a) Fatal

(b) Serious injury

(c) Minor injury

(d) Possible Injury

(e) No injury/PDO

**Figure 5 The SHAP values, explaining the contribution of EVs to the raw TabNet model output for a specific observation.**

## Conclusion

In the realm of transportation safety, understanding pedestrian crash severity is crucial, particularly due to the inherent vulnerability of pedestrians. This study focused on employing TabNet, an advanced deep learning (DL) method designed for tabular data. Our use of SHAP techniques for interpretation further enhanced our understanding of TabNet's application. The findings indicated that TabNet was exceptionally effective in analyzing pedestrian crash data from Utah. However, employing TabNet did pose challenges, particularly in hyperparameter tuning and model interpretation. For instance, tuning hyperparameters for TabNet required a considerable amount of time — 20 hours and 16 minutes — on a general computer setup (Core i7- 9th generation with 32 GB RAM). Moreover, interpreting the TabNet results using SHAP was a time-intensive process, taking approximately 68 hours and 31 minutes. This highlights a crucial trade-off: the choice between achieving high accuracy with DL and ML models, which necessitates more time, versus opting for faster but potentially less accurate results from statistical methods.

In summary, our study provides valuable insights for transportation engineers in choosing appropriate methods for analyzing pedestrian crash severity. The methodologies and approaches we employed, especially focusing on TabNet, offer a framework that can be adapted for broader crash variable investigations in the field of transportation safety.